\documentclass{article}

\usepackage[english]{babel}

\usepackage[letterpaper,top=2cm,bottom=2cm,left=3cm,right=3cm,marginparwidth=1.75cm]{geometry}

\usepackage{amsmath}
\usepackage{graphicx}
\usepackage[colorlinks=true, allcolors=blue]{hyperref}

\usepackage{graphicx}
\usepackage[table]{xcolor}
\usepackage{booktabs}
\newcolumntype{K}{!{\color{white}\ }c}
 \usepackage{array,multirow}

\usepackage{tablefootnote}
\usepackage{listings}
\definecolor{dkblue}{rgb}{0,0,.6}

\title{\bf \textsc{Recover}: A Neuro-Symbolic Framework for \\ Failure Detection and Recovery}

\author{Cristina Cornelio\footnote{Samsung AI, Cambridge, UK {\tt\small c.cornelio@samsung.com}} 
~~~ Mohammed Diab\footnote{Mohammed Diab, University of Plymouth, UK {\tt\small mohammed.diab@plymouth.ac.uk}}}

\date{}

\begin{document}
\maketitle

\begin{abstract}

Recognizing failures during task execution and implementing recovery procedures is challenging in robotics. 
Traditional approaches rely on the availability of 
extensive data 
or a tight set of constraints, while more recent approaches leverage large language models (LLMs) to verify task steps and replan accordingly.
However, these methods often operate offline, necessitating scene resets and incurring in high costs.
This paper introduces \textsc{Recover}, a neuro-symbolic framework for online failure identification and recovery. 
By integrating ontologies, logical rules, and LLM-based planners, \textsc{Recover} exploits symbolic information to  enhance the ability of LLMs to generate recovery plans and also to decrease the associated costs.
In order to demonstrate the capabilities of our method in a simulated kitchen environment, we introduce \textsc{OntoThor}, an ontology describing the AI2Thor simulator setting.
Empirical evaluation shows that \textsc{OntoThor}'s logical rules accurately detect all failures in the analyzed tasks, and that \textsc{Recover} considerably outperforms, for both failure detection and recovery, a baseline method reliant solely on LLMs. 
Supplementary material is available at: \url{https://recover-ontothor.github.io}.
\end{abstract}

%%%%%%%%%%%%%%%%%%%%%%%%%%%%%%%%%%%%%%%%%%%%%%%%%%%%%%%%%%%%%%%%%%%
\section{Introduction}

With the increasing use of robots in tasks involving humans in the perception-action loop, understanding the reasons behind failures in both planning and execution is a significant challenge for enhancing the reliability, adaptability, and safety of autonomous systems. Robots need to comprehend why and when failures occur and devise appropriate solutions based on the current situation. To achieve this, robots should be equipped with robust planning, perception, and reasoning capabilities enabling them to analyze failures and propose recovery strategies in real time.

The standard approaches to autonomous robots are typically model-based or policy-based~\cite{chatzilygeroudis2019survey}. Model-based approaches can involve offline planning, where the robot considers the current state and utilizes its model to predict the next state and potential rewards, enabling it to plan a sequence of actions expected to maximize reward. In online model-based planning instead, the robot continuously re-plans based on the current state, adjusting its actions in response to changes in the environment. 
Policy-based approaches usually entail either open-loop policy, where the robot predicts a sequence of actions based on the initial state and goal, or closed-loop policy, where the robot predicts individual actions at each moment based on the current state and goal. These policies guide the robot's decision-making process, facilitating adaptive behavior in dynamic environments.
In our approach, we integrate elements from both methodologies mentioned above: we emply an ontology and a set of rules to represent the environment, and utilize these representations to refine a policy determined by a large language model (LLM) acting as a planner. This combined strategy enables the model to guide the policy, especially in scenarios where limited data availability restricts the policy's exposure to relevant instances.

In this paper we introduce an innovative use of ontological knowledge bases and LLMs for failure recognition and recovery within robotic systems, thereby enhancing the overall reliability and efficiency of robotic task execution. 
Our framework, named \textsc{Recover}, leverages the available symbolic knowledge about an environment (e.g., the set of available objects and their properties) to efficiently detect failures when they occur, in an online fashion {\it during} the task execution. The symbolic representation, embedded within an ontology, enables the robot to map multi-modal data (e.g., video, images, and audio) to the same representation, allowing simultaneous reasoning across all data types.
When a failure has been identified, an LLM is employed as a re-planner, producing a set of steps to perform to recover from the failure and complete the task. 

Employing LLMs for planning is not novel and has been recently proposed in various contexts~\cite{chain_of_thoughts,ahn2022can,yao2020calm,huang2022language,schick2023toolformer}. However, LLMs come with several limitations, including a tendency to ``hallucinate''~\cite{golovneva2022roscoe,pan2024unifying}.
By incorporating symbolic information, we can steer the LLM-planner towards generating fewer hallucinations, thereby ensuring that the system operates within the confines of available objects and actions.

Current approaches utilizing LLMs for failure recovery and re-planning typically function in an offline manner: initially, a plan is executed in full to solve a task; if a failure is detected, a revised plan is generated and the task execution restarts from the initial state. This necessitates resetting the scene before executing the revised plan. 
Performing this method iteratively enables the refinement of the plan until one that successfully leads to task completion is generated~\cite{wang2023learning}. However, this approach is impractical in real-world scenarios where, for instance, the environment undergoes changes during task execution (such as an object being broken or moved), or when, as often happens in robotics, even a correct plan may result in failure during execution (such as when a robot struggles to perform a feasible action).
In such scenarios, the ability to act in real-time is crucial.
Our framework detects failures during task execution and generates a new plan based on the environment conditions at the moment of failure, without needing to observe the effects of the failure throughout the entire plan.

Utilizing symbolic knowledge alongside natural language also confers explainability features to the framework, enabling transparent and understandable reasoning behind the robot's actions and responses to failures. This attribute not only enhances trust and transparency but also helps in debugging and optimizing the system.

Furthermore, the use of ontologies for storing the environmental description and reasoning enables the integration of specialized knowledge and personalization, thereby adapting the system's capabilities to particular tasks or environments, enhancing its effectiveness and versatility. 
As a contribution of our work, we extend established taxonomies~\cite{diab2019ontology} of failure categories in a human-robot interaction context, incorporating preferences such as allergies and dietary restrictions. 
We release an ontology named \textsc{OntoThor}, specifically designed for the AI2Thor simulator, which we utilize in our experiments. \textsc{OntoThor} describes kitchen environments and includes personalized features. 

This kind of ontologies are particularly useful when interacting with human agents who may possess diverse preferences or requirements, or in scenarios involving technical and specialized equipment. 
Examples of such use cases are healthcare support (e.g., surgical robots or assistant bots), automated transportation (including spacecraft, planes, or cars), and household assistance (e.g., kitchen robots or cooking assistants).

\subsection{State of the Art}

Recent studies have investigated the utilization of pre-trained LLMs for planning and executing actions in interactive environments, by using the priors of the LLM for plan generation~\cite{huang2022language,wu2023tidybot}.
Usually, this involves converting multi-modal observations into natural language, utilizing an LLM to generate domain-specific actions or plans, and then employing an agent for execution. However, these approaches are susceptible to hallucination and lack deep understanding and reasoning capabilities~\cite{ahn2022can}.

Reasoning, especially when grounded on concrete actions, is one of the most desired capability of LLMs, as it offers explainability and enables the generation of more meaningful solutions~\cite{chain_of_thoughts,narangself}.
However, despite substantial progress in the reasoning abilities of LLMs~\cite{wei2022emergent}, the requirement to detect and correct errors in the reasoning process remains an open problem.

Some progress has been made in this regards:
ReAct~\cite{yao2022react} is a prompting refinement technique that leverages the reasoning and planning capabilities of LLMs to iteratively refine a plan to solve a task. Although it introduces innovative features like iterative action-environment observations and LLM-policy refinement, it may face challenges in complex or real-world environments.  
Reflexion~\cite{shinn2023reflexion} offers a unique approach to reinforcing LLMs-based agents through verbal reflection on task feedback. By maintaining reflective text in an episodic long-term and short-term memory, it enhances decision-making in subsequent trials across diverse tasks.  
As an extension of Reflexion, ExpeL~\cite{zhao2023expel}, SALAM~\cite{wang2023learning}, and SAMA~\cite{li2023semantically} enhance their capabilities by analyzing patterns in failures and successes to extract valuable insights. These insights are then integrated into prompts, thereby improving decision-making and adaptive behavior. 
RetroFormer~\cite{yao2024retroformer} introduces a framework to bolster LLMs-based agents by refining prompts based on environment feedback. Through policy gradient techniques, the system adjusts prompts iteratively, learning from rewards across various tasks and environments. This process enhances the LLM's performance by summarizing past failures and suggesting action plans for improvement.

The increasing use and advancements of LLMs in recent years have led to their widespread application across various research fields in AI, including robotics \cite{ren2023robots,das2021semantic,jin2024robotgpt,vemprala2023chatgpt}.
Most relevant to our work is REFLECT~\cite{liu2023reflect}  a framework that, similarly to Reflexion, utilizes LLMs in the context of  robot failures to explain them and propose corrective plans. 
REFLECT demonstrates great promise in addressing robot failures and proposing corrective plans, however, it is currently unable to accommodate human preferences and personalization.
Moreover, the necessity to observe the entire plan, reset the scene, and the cost of the extensive use of LLMs (both computational and monetary) are strong limitations.

%%%%%%%%%%%%%%%%%%%%%%%%%%%%%%%%%%%%%%%%%%%%%%%%%%%%%%%%%%%%%%%%%%%
\section{The Method: Ontology-based Failure Identification and Recovery}\label{sec:method}

\begin{figure}[h]
  \centering
  \includegraphics[width=\linewidth]{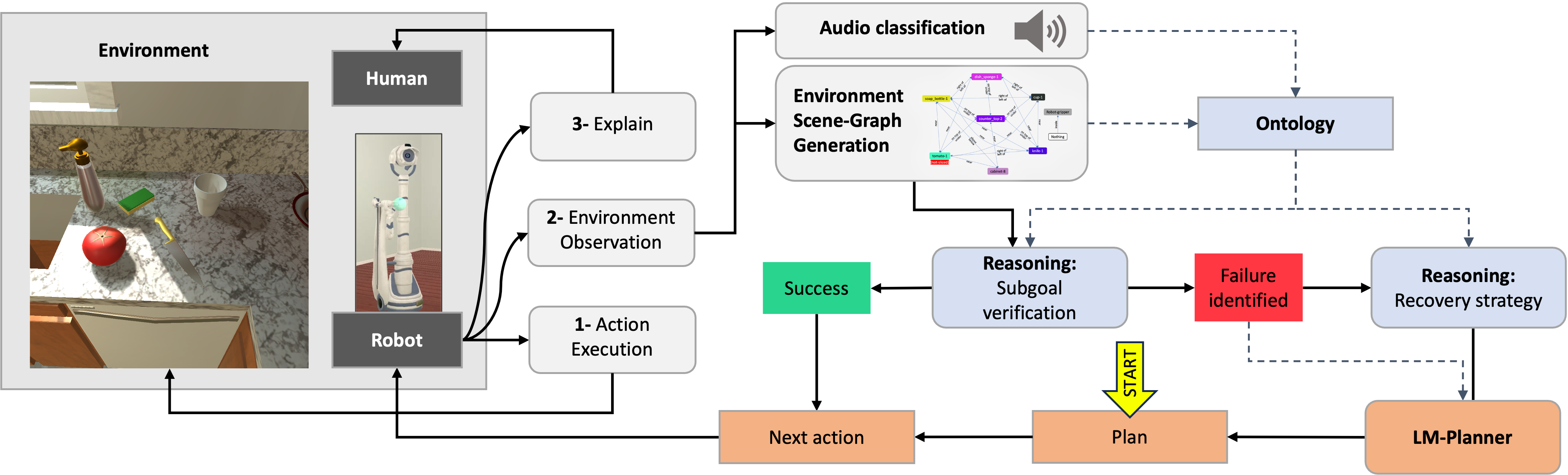}
  \caption{Overview of the \textsc{Recover} framework. Starting with a plan the robot executes one action at a time over the environment. The outcome of each action conveyed through multi-modality information (audio and video) is then processed and converted into an audio label and a scene-graph. These are stored within the ontology and provided as input to the sub-goal verifier, which classifies the action execution as either a failure or a success. If the action is successful, the robot proceeds with the next step. In the event of failure, reasoning module will use the failure information to extract the recovery strategy from the ontology and will supply it to the LLM-planner.
  Subsequently, the LLM-planner generates a new plan to recover from the failure and accomplish the task.
  \emph{Figure legend:} Blue elements represent model-based components, while orange elements denote policy-based components. Sharp-cornered shapes indicate input/output elements, whereas round-cornered shapes signify procedures. Dashed lines correspond to input sources, while solid lines indicate the procedural loop.}
  \label{fig:pipeline_overview}
\end{figure}

\textsc{Recover} is a neuro-symbolic framework for failure identification and recovery that exploits symbolic knowledge to aid the execution of a task. The symbolic knowledge is provided in the form of an ontology and a set of logical rules that describe the environment, the actions of the robot, and the preferences of human agents. Moreover, it contains information about the possible failures that can occur in that particular environment and the corresponding recovery strategies. An overview of the \textsc{Recover} methodology is depicted in Figure~\ref{fig:pipeline_overview}.

To solve an input task, \textsc{Recover} takes as input a plan to execute and an ontology describing the environment.
The plan comprises a symbolic, however interpretable, sequence of actions that the robot can perform on a given environment (e.g., {\it pick\_up(mug)}). 
An ontology is a formal description of a domain or an environment, organizing concepts, classes, and entities along with the relationships between them. 
The knowledge is organized in the form of unary (representing the type or class of an entity) and binary (representing the relationship that might exist between two entities) predicates.
Moreover, an ontology also has a set of rules that define properties and relations between the classes (e.g., all the objects of class Mug are also Fillable objects).

\newpage
\noindent
In our scenario, the ontology describes the environment in which the robot is acting, the preferences of one or more human agents (e.g., dietary restrictions), and safety conditions (e.g., the stove must not be left on). During task execution, the ontology is populated by entities (also called instances), such as concrete objects, agents or events (e.g., mug-1, human-3, event\_002). In what follows we will denote the ontological classes starting with a capital letter (e.g., class Apple) and instances/entities with lower case letters (e.g., object apple-1).

The framework starts with the execution of the first action in the plan. The action is given to the robot that will (1) execute the action on the environment; (2) observe the status of the environment after the action has been performed; and (3) explain the action to the human using an LLM to translate the step into natural language.

While the action is being executed, the robot records any sounds that may occur, and after the action is completed, the robot observes the scene.
The auditory information gathered by the robot is mapped by a classifier into a label $s$ chosen within a set of available sound classes $\{s_1,\ldots,s_k\}$ (e.g., {\it ToggleOnMicrowaveSound} or {\it CloseFridgeSound}) and then stored in the ontology in the form of triples (e.g., (\textit{event\_001, has\_sound, s}) and (\textit{s, type, CloseFridgeSound})).

The visual information gathered by the robot after the action execution is processed into a scene-graph. 
A scene-graph is a graph where each node corresponds to an element in an image. Each node is classified to belong to a certain class $c \in \{c_1, \ldots, c_n\}$  (e.g., Apple) representing the type information of the object. The edges between the nodes correspond to the binary relation $r$ between two objects in the scene, and is chosen within a set of possible spatial relations $\{r_1,\ldots,r_m\}$. Each edge and the two corresponding nodes constitute a triple (e.g., (\emph{apple-1}, \emph{on-top-of}, \emph{table-1})). Additionally each object can have one or multiple states, that represent the condition of the object at a particular time step (e.g., a mug can be dirty or clean). 
\begin{figure}[h]
  \centering
  \includegraphics[width=\linewidth]{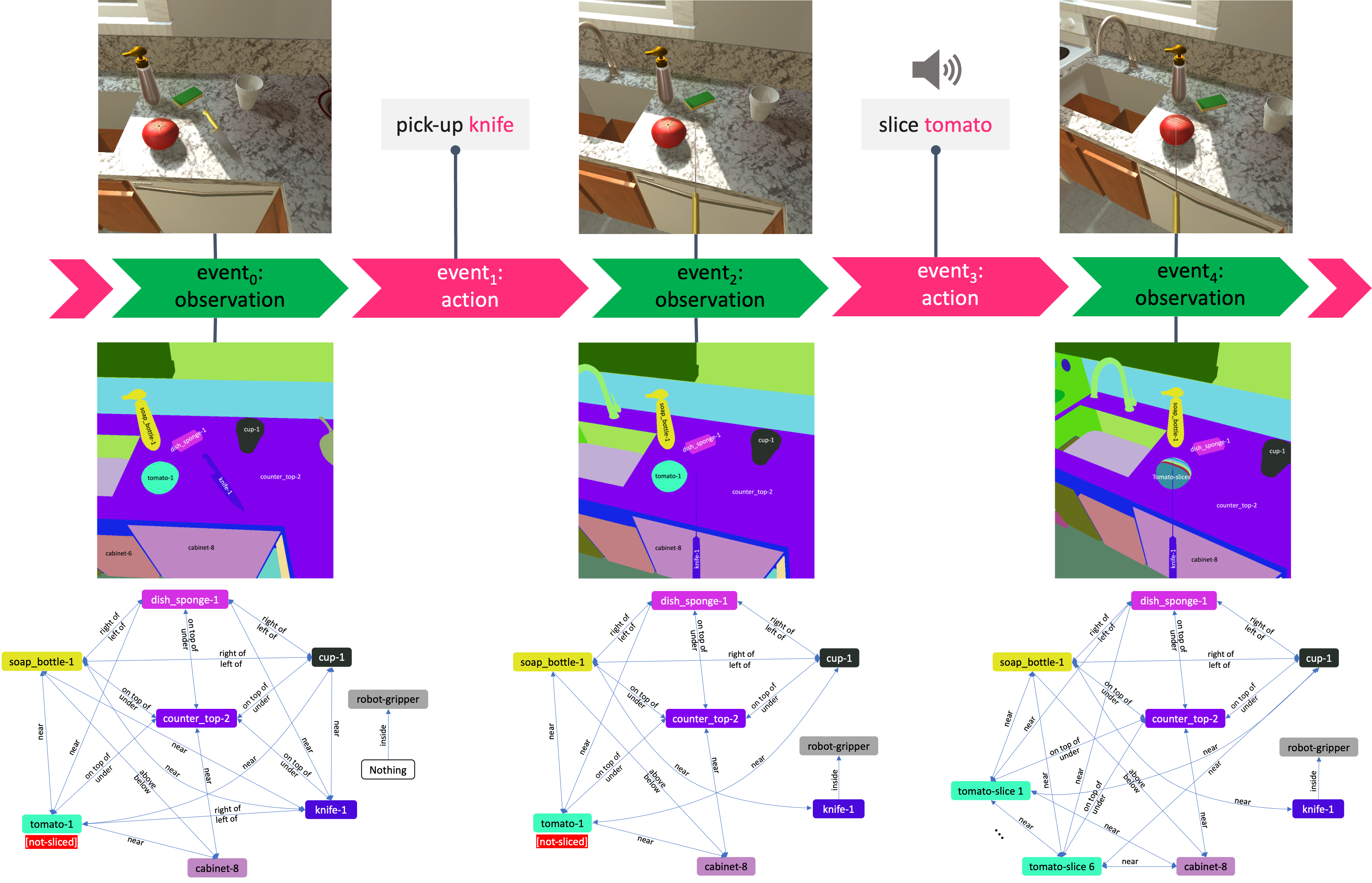}
  \caption{Example of the information flow during plan execution, depicting the alternation of two event types: observation events (in green) and action events (in pink). Each action event can be associated with a sound recorded by the robot during the action's execution. Each observation event is linked to a scene-graph describing the frame captured by the robot. In the scene graph, each node represent an object identified in the scene and each edge represent the relation between two objects. The colors of the nodes in the scene graph correspond to those in the segmentation image (the adjacent image) where the objects have been identified.}
  \label{fig:pipeline_example}
\end{figure}

Figure~\ref{fig:pipeline_example} shows three scene-graphs corresponding to the scenes observed by the robot before and after two action executions.

Once the environment observation results are stored in symbolic form within the ontological knowledge base, a set of logical rules, defining potential failures in the environment are applied. The application of the rules is done with a symbolic reasoner, such as a SPARQL reasoning engine, a RDF-reasoning engine (e.g., RDFOX~\cite{nenov2015rdfox}) or a more expressive reasoner (e.g., prolog~\cite{swi_prolog} for first order logic). The result of the reasoning process is provided as either a success state, meaning that the action was performed successfully, or as a failure instance (e.g., {\it DroppingObjFailure}), meaning that the action was not performed successfully. In a success state, the robot will proceed with the next step of the plan. Instead, if a failure is inferred by the rules, the robot needs to replan before proceeding with the next action. 

A set of recovery instructions are stored in the ontology for each failure type. These are provided as input to the LLM, along with the original plan, the goal of the task, the success condition, and the environment state in the form of natural language text. 

The LLM then generates the new plan as a sequence of steps. This text is then mapped to a sequence of commands executable by the robot by computing a similarity measure between the embedding of each sentence and the elements in the pool of available actions that the robot can execute in a given environment.
Once the new plan is finalized, the robot resumes the loop by executing the next available action.

%%%%%%%%%%%%%%%%%%%%%%%%%%%%%%%%%%%%%%%%%%%%%%%%%%%%%%%%%%%%%%%%%%%

\section{OntoThor Ontology}

We created an ontology, named \textsc{OntoThor}, characterizing the kitchen environment of AI2Thor simulator~\cite{ai2thor}.
AI2Thor is a 3D simulator designed for research in embodied artificial intelligence. It provides a realistic 3D environment where agents can navigate and interact with objects in a virtual home setting.  AI2Thor offers various functionalities such as object manipulation, scene understanding, and task execution and it serves as a valuable tool for studying navigation, object interaction, and spatial understanding within indoor environments.

\textsc{OntoThor} contains the following classes describing the environment:
\begin{itemize}
\item {\bf{ Action}}: Covers agent behaviors and is categorized into subclasses based on interactions with objects and surroundings. Includes actions with held objects (e.g., crack, pour), without (e.g., open, close), object movement (e.g., pickup, place), and non-interactive actions (e.g., observation). Facilitates understanding and reasoning about kitchen behaviors.
\item {\bf Agent}: Includes humans and robots. Humans are divided in classes by preferences (e.g., Celiac, Vegan etc.) aiding customized interactions and decision-making based on individual needs.
\item {\bf PhysicalObject}: Categorizes kitchen items into groups such as consumables (e.g., fruits, vegetables), cookware (e.g., pots, pans), and appliances (e.g., ovens, coffee machines). This organization simplifies the process of identifying, understanding, and using kitchen items effectively.
\item {\bf PhysicalProperty}: Defines attributes like breakable, fillable, and receptacle for kitchen objects, ensuring appropriate use that reflects the functionalities. For example, ceramic plates are breakable, and pots are fillable.
\item {\bf Sound}: Categorizes sounds by their source and interaction: \emph{AppliancesSound} like opening fridges and toggling faucets, \emph{DroppingSound} dropping or breaking objects, and \emph{ObjectInteractionSound} for sounds like slicing food or pouring liquids. This ensures a clear, organized representation of auditory information in kitchen activities.
\item {\bf SpatialRelation}: Defines positional relationships between objects, like above, under, on top of, to the right of, to the left of, inside, near, and blocking, aiding understanding of spatial configurations.
\item {\bf State}: Describes the current condition of physical objects in the environment at a specific point in time, providing temporal tracking of object conditions. For instance, it can indicate that a cup was full of liquid at a particular time step.
\item {\bf Time}: Includes discrete time for tracking events at discrete time steps or continuous time stamps for precise event duration. This enables accurate temporal modeling of events.
\item {\bf Location}: Categorizes areas such as bathroom, bedroom, kitchen, and living room, enabling precise identification and contextual reasoning regarding physical object functionality within these spaces. In this work, we focused on the kitchen environment only.
\end{itemize}

Once the task execution is started, the ontology is populated by all the entities/individuals present and/or discovered during the task execution and by all the events that occurred.
For example, if an object of class \textit{Apple} is present in the scene, an entity called \textit{apple-1} is created under that class. A number is always part of each object instance name to uniquely identify the object in the room. This is necessary since there might be multiple instances of the same object type (e.g. there are 2 apples in the kitchen). In what follows, for simplicity, we will omit the numbers when not necessary.

We defined the following classes in the ontology to facilitate the storage of events during the robot's task execution:
\begin{itemize}
\item {\bf Event}: Describes both observation and action events, as described in Figure~\ref{fig:pipeline_example}. \emph{Action events} detail agent actions, including the involved source/target object, timing information and presence and type of sound that occurred during the action execution. \emph{Observation events} record the states of physical objects and their spatial relations at a specific time in the form of scene-graph triples.
\item {\bf Triple}: Employs second-order logic to describe the state of the environment. Each triple contains a subject, predicate, and object such as (\emph{mug-1}, \emph{near}, \emph{cabinet-7}). These triples are linked only to observation events using the \emph{hasTriple} relationship, enabling a comprehensive semantic description of the current event. 
%This structured approach allows for the representation of complex relationships between entities within the environment.
\item {\bf RecoveryStrategy}: Guides the robot's actions in specific failure situations, considering factors such as object state and material. For example, if a cup breaks, the strategy may involve selecting a new mug; however, if it's plastic, the robot may opt instead to retrieve it. These strategies enhance the robot's adaptive decision-making capabilities.
\end{itemize}

As described in Section~\ref{sec:method} (see also Figure~\ref{fig:pipeline_example}), during the task execution there is an alternation of {\it action events} and {\it observation events}. 
Thus, for each step in the plan, the ontology creates 2 events: $\text{\it event}_i$ with an associated action (e.g., \emph{pick\_up}) and $\text{\it event}_{i+1}$ with an associated observation-action. For example, in Figure~\ref{fig:pipeline_example}, there are 5 events: 
$\text{\it event}_0$, $\text{\it event}_2$, and $\text{\it event}_4$ with action \emph{observation}; 
$\text{\it event}_1$ with action \emph{pick\_up} and target object \emph{knife}; and 
$\text{\it event}_1$ with action \emph{slice} and target object \emph{tomato}.

Each action event might also have sound information: for example in Figure~\ref{fig:pipeline_example} the event $\text{\it event}_1$ is not associated with any sound while  $\text{\it event}_3$ (with action {\it slice}) has sound $\text{\it sound}_3$ of type \emph{SliceVeggySound}.  

As mentioned above, each observation event is connected to a set of triples that represent the scene-graph information. In Figure~\ref{fig:pipeline_example} the scene graph associated with the observation actions is depicted by the graph in the the bottom row of the image. For example, $\text{\it event}_2$ has 14 triples including (\emph{knife-1}, \emph{inside}, \emph{robot-gripper}), (\emph{tomato-1}, \emph{near}, \emph{soap-bottle-1}), and (\emph{dish-sponge-1}, \emph{on-top-of}, \emph{counter\_top-2}).

This approach permits the storage of the entire process within the ontology, allowing the utilization of reasoning to detect the presence of any failures.

\begin{figure}
  \centering
  \includegraphics[width=0.8\linewidth]{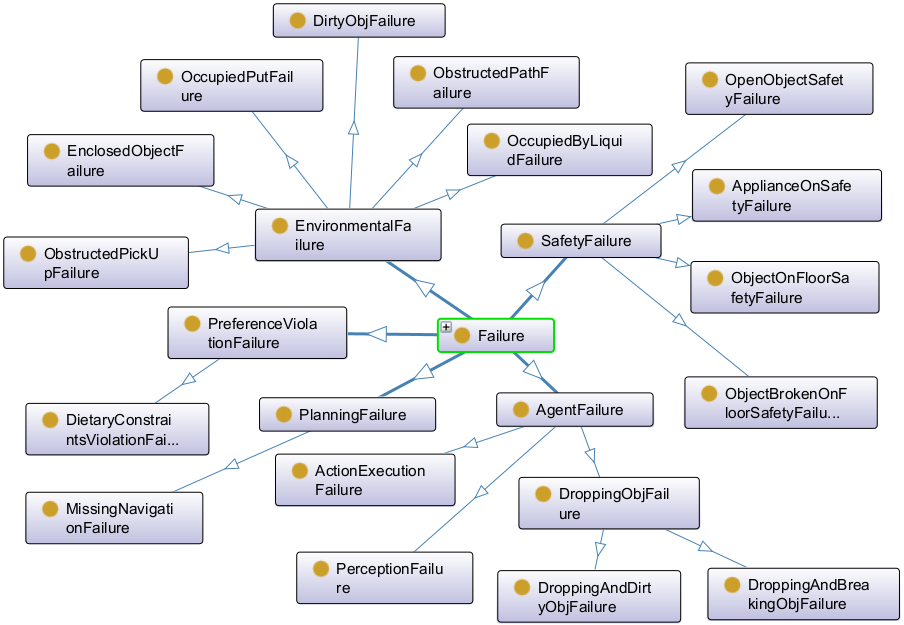}
  \caption{Failures taxonomy in \textsc{OntoThor}} %(figure generated using done using Protege)}
  \label{fig:failure_taxonomy}
\end{figure}

\subsection{Failures Classification and Detection Rules}
The ontology categorizes also the potential types of failures that may arise in the AI2Thor environment, as depicted in Figure~\ref{fig:failure_taxonomy}. 
The class {\bf Failure} is a subclass of the {\bf Event} class, as a failure can potentially be identified within each event.
The classification is organized as follows:
\begin{itemize}
\item {\bf AgentFailure}: Comprises diverse sub-classes representing different failure types during operation, including action execution failure, dropping object failure (potentially leading to breakage or soiling), and perception failure (e.g., misidentifying objects).
\item {\bf EnvironmentalFailure}: Contains various types of failures related to environmental conditions. It covers situations such as dirty objects, inaccessible or obstructed paths, and receptacles filled with liquid or occupied placement regions.
\item {\bf PlanningFailure}: Addresses situations such as missing or wrong steps in the plan, to ensure effective task planning.
\item {\bf PreferenceViolationFailure}: Is employed to recognize personalization failures related to human preferences. For instance, it enables the identification of situations where individuals dislike specific ingredients or adhere to a specific diet regimen.
\item {\bf SafetyFailure}: Is used to maintain a secure environment. For instance, if a glass breaks and shards are scattered on the floor, this would be classified as a safety failure.
\end{itemize}

Additionally, we defined a set of rules, each corresponding to a specific failure type that may arise within the AI2Thor kitchen environment. 
The rules are applied following each observation event to identify whether the executed action has resulted in a failure or not.
For example, a rule to identify a failure generated by dropping an object ({\it DroppingObjFailure}) is as follows:
\begin{lstlisting}[escapeinside={(*}{*)}]
Event(e) (*$\wedge$*) hasAction(e,a)  
(*$\wedge$*) (ActionWithHeldObject(a) (*$\vee$*) NonInteractiveAction(a)) 
(*$\wedge$*) hasPreconditions(e,pre_c) (*$\wedge$*) hasTriple(pre_c,trp1) 
(*$\wedge$*) hasSubject(trp1,held_obj1) (*$\wedge$*) (*$\neg$*)(Nothing(held_obj1)) 
(*$\wedge$*) hasObject(trp1,rg) (*$\wedge$*) RobotGripper(rg)  
(*$\wedge$*) hasPostconditions(e,post_c) (*$\wedge$*) hasTriple(post_c,trp2)
(*$\wedge$*) hasSubject(trp2,held_obj2) (*$\wedge$*) Nothing(held_obj2)
(*$\wedge$*) hasObject(trp1,rg) (*~~~~~~~~$\rightarrow$*) DroppingObjFailure(e)
\end{lstlisting}
where the symbols $\wedge$, $\vee$ and $\neg$ are the standard AND, OR and NOT logic operators and $\rightarrow$ corresponds to the logic implication.
This rule states that if the robot was holding an object (e.g., a knife) and performs an action with it (e.g., cutting an apple) and after the action it is not holding the object anymore, it means that the robot dropped the object during the execution of the action.
More formally, the rule states that if there is an event $e$ where 
(lines 1 and 2) the action $a$ to be performed is not placing/picking-up an object or an empty-gripper action;
(lines 3, 4 and 5) the robot was holding an object $held\_obj1$ before the action (since there is a triple $trp1$ in the preconditions  $pre\_c$ -- the environment status before the action execution -- that indicate that something is in the robot gripper); and
(lines 6, 7 and 8) the robot is not holding the object anymore after the action (since there is a triple $trp1$ in the post-conditions $post\_c$ -- the environment status after the action execution -- that indicate that nothing is in the robot gripper);
then the event $e$ is an instance of a \emph{DroppingObjectFailure}.

%%%%%%%%%%%%%%%%%%%%%%%%%%%%%%%%%%%%%%%%%%%%%%%%%%%%%%%%%%%%%%%%%%%

\section{Experimental setting}

\begin{table}[t]
  \centering
  \begin{tabular}{lc l c c}
    \toprule
    & ID & Name & \#steps & \#objects \\
    \midrule
\multirow{7}{*}{\rotatebox[origin=c]{90}{easy}}
& T1 & Serve wine & 8 & 2\\ 
& T2 & Make coffee & 9 & 2\\ 
& T3  & Boil water in a pot & 10 & 2\\ 
& T4 & Fry egg in a pan & 11 & 2\\ 
& T5 & Toast bread & 12 & 3\\ 
& T6 & Warm water (in microwave) & 16 & 3\\ 
& T7 & Cook potato slice (in microwave) & 20 & 3\\ 
\midrule
\multirow{5}{*}{\rotatebox[origin=c]{90}{complex}}
& T8 & Simple salad & 26 & 4\\ 
& T9 & Clean/order kitchen & 28 & 7\\ 
& T10 & Vegetarian sandwich & 30 & 5\\ 
& T11 & Cook egg and potato slice & 32 & 7\\ 
& T12 & Complex salad & 33 & 7\\ 
\bottomrule  
  \end{tabular}
    \caption{Tasks implemented in the experiments}
  \label{tab:tasks}
\end{table}

For the experiments we implemented 12 tasks and 12 failure types in the kitchen scenario of the AI2THor simulator~\cite{ai2thor}. An overview of the different tasks is provided in Table~\ref{tab:tasks}. The tasks are ordered based on the number of steps that are present in the original plan. Moreover, the tasks are divided in two categories: easy tasks with at most 20 steps, and complex tasks that have more than 20 steps and more than 3 objects. 

The failures implemented in the experiments are a subset of the ones described in the ontology (due to the limitations of the simulator):
(1) \emph{EnclosedObjectFailure};
(2) \emph{DroppingObjFailure};
(3) \emph{DroppingAndDirtyObjFailure};
(4) \emph{DroppingAndBreakingObjFailure};
(5) \emph{DirtyObjFailure};
(6) \emph{OccupiedPutFailure};
(7) \emph{PlanningFailure} (missing step);
(8) \emph{ActionExecutionFailure};
(9) \emph{DietaryConstraintsViolationFailure};
(10) \emph{PlanningFailure} (wrong step);
(11) \emph{OccupiedByLiquidFailure}; and 
(12) \emph{MissingNavigationFailure}.
Not all the failures are possible in all the tasks: for example \emph{DroppingAndBreakingObjFailure} is not available for the task ``Boil water in a pot'' since a pot is the only object that is picked up during the task and it is a non-breakable object. All the failure-task combinations that are not available are indicated with a gray box in Table~\ref{tab:results}.

\subsection{Implementation details}

We implemented the general pipeline in an entirely modular fashion, such that each component can be easily replaced with alternative methods. 

\smallskip\noindent
{\bf Planning.} We used the ground-truth original plan for each task, while the recovery plan is generated by using GPT-4 as LLM-based planner. Alternatively, the original plan can be automatically generated using an LLM as well.
The output of the LLM is mapped to the set of feasible actions using a large pre-trained sentence embedding model (similarly to \cite{liu2023reflect,huang2022language}).

\smallskip\noindent
{\bf Scene graph generation.} Techniques to transform an image into a scene-graph have drastically improved in the last years. Scene graph generation can be done in different ways: for example by layering different models such as image segmentation, image classification and scene-graph predicate classification (as in~\cite{liu2023reflect}), or with an end-end model (a popular example is the work of Knyazev et al.~\cite{sgg}). 
In our experiments we follow the approach of Liu et al.~\cite{liu2023reflect}. This is a rule-based approach for predicate classification that, given the ground-truth bounding boxes, labels and state of the objects present in a scene, computes the relation label between each pair of objects based on the proximity and relative location of the items.

\smallskip\noindent
{\bf Sound classification.} In the experiments we use the ground-truth labels for the detected sounds. An alternative neural-based sound classifier (such as \cite{guzhov2022audioclip}) can be used and is available in our code.

\smallskip\noindent
{\bf Failure detection.} The failure rules are written in SPARQL, and their application is done with the SPARQL engine available in OwlReady2 python library~\cite{lamy2017owlready}. We use the same library to store and query the ontology.

The experiments were performed on a Linux machine with 20 cores and an NVIDIA GeFOrce RTX 3090Ti GPU.

In the experiments we assume that there is at most one failure per task. A straightforward generalization of \textsc{Recover} would allow the identification and correction of nested failures (failures that happen during the recovery plan) as well.

%%%%%%%%%%%%%%%%%%%%%%%%%%%%%%%%%%%%%%%%%%%%%%%%%%%%%%%%%%%%%%%%%%%

\section{Experimental Results}

\begin{table}[t]
  
  \centering
  \begin{tabular}{@{\hskip 0em}c@{\hskip 0.5em}c@{\hskip 0.3em}KKKKKKKKKKKK}
    \toprule
    & \multicolumn{12}{c}{Failures} \\
    \cmidrule{3-14}
    &  & \phantom{1}1 & \phantom{1}2 & \phantom{1}3 & \phantom{1}4 & \phantom{1}5 & \phantom{1}6 & \phantom{1}7 & \phantom{1}8 & \phantom{1}9 & 10 & 11 & 12  \\ 
    \midrule
    
    \addlinespace[.3em]
    \multirow{8}{*}{\rotatebox[origin=c]{90}{easy tasks}} &
    T1 
    & $\ast$\cellcolor{teal!60} & \cellcolor{lightgray!30} & \cellcolor{lightgray!30} & \cellcolor{lightgray!30} & \cellcolor{red!40} & \cellcolor{lightgray!30} & \cellcolor{teal!60} & \cellcolor{red!40} & \cellcolor{lightgray!30} & \cellcolor{red!40} & \cellcolor{red!40} & $\ast$\cellcolor{teal!60} 
    \\\addlinespace[.25em]
    & T2  
    & \cellcolor{teal!60} & \cellcolor{lightgray!30} & \cellcolor{lightgray!30} & \cellcolor{teal!60} & \cellcolor{red!40} & \cellcolor{lightgray!30} & \cellcolor{red!40} & \cellcolor{teal!60} & \cellcolor{lightgray!30} & \cellcolor{red!40} & \cellcolor{red!40} & \cellcolor{teal!60}
    \\\addlinespace[.25em]
    & T3  
    & \cellcolor{lightgray!30} & \cellcolor{lightgray!30} & \cellcolor{teal!60} & \cellcolor{lightgray!30} & \cellcolor{lightgray!30} & \cellcolor{teal!60} & \cellcolor{red!40} & \cellcolor{teal!60} & \cellcolor{lightgray!30} & \cellcolor{teal!60} & \cellcolor{lightgray!30} & \cellcolor{teal!60}
    \\\addlinespace[.25em]
    & T4 
    & \cellcolor{teal!60} & \cellcolor{lightgray!30} & \cellcolor{lightgray!30} & \cellcolor{teal!60} & \cellcolor{teal!60} & $\ast$\cellcolor{teal!60} & \cellcolor{teal!60} & \cellcolor{red!40} & \cellcolor{lightgray!30} & \cellcolor{teal!60} & \cellcolor{lightgray!30} & \cellcolor{teal!60}
    \\\addlinespace[.25em]
    & T5  
    & \cellcolor{teal!60} & \cellcolor{teal!60} & \cellcolor{lightgray!30} & \cellcolor{lightgray!30} & \cellcolor{lightgray!30} & \cellcolor{lightgray!30} & \cellcolor{red!40} & \cellcolor{teal!60} & \cellcolor{lightgray!30} & \cellcolor{red!40} & \cellcolor{lightgray!30} & \cellcolor{teal!60}
    \\\addlinespace[.25em]
    & T6  
    & \cellcolor{teal!60} & \cellcolor{lightgray!30} & 
    \cellcolor{lightgray!30} & \cellcolor{teal!60} & 
    \cellcolor{lightgray!30} & \cellcolor{teal!60} & 
    \cellcolor{red!40} & $\ast$\cellcolor{teal!60} & 
    \cellcolor{lightgray!30} & \cellcolor{teal!60} & 
    \cellcolor{lightgray!30} & \cellcolor{teal!60}
    \\\addlinespace[.25em]
    & T7  
    & \cellcolor{teal!60} & \cellcolor{teal!60} & \cellcolor{lightgray!30} & \cellcolor{red!40} & \cellcolor{red!40} & \cellcolor{teal!60} & \cellcolor{red!40} & \cellcolor{teal!60} & \cellcolor{lightgray!30} & $\ast$\cellcolor{teal!60} & \cellcolor{lightgray!30} & \cellcolor{teal!60}
    \\\addlinespace[.25em]
    
    \midrule
    
    \addlinespace[.3em]
    
    \multirow{6}{*}{\rotatebox[origin=c]{90}{complex tasks}}
    & T8 
    & $\ast$\cellcolor{teal!60} & $\ast$\cellcolor{teal!60} & 
    \cellcolor{lightgray!30} & \cellcolor{lightgray!30} & 
    \cellcolor{red!40} & \cellcolor{lightgray!30} & 
    $\ast$\cellcolor{teal!60} & $\ast$\cellcolor{teal!60} & 
    \cellcolor{teal!60} & $\ast$\cellcolor{teal!60} & 
    \cellcolor{red!40} & $\ast$\cellcolor{teal!60}
    \\\addlinespace[.25em]
    & T9   
    & \cellcolor{lightgray!30} & \cellcolor{teal!60} & \cellcolor{lightgray!30} & \cellcolor{lightgray!30} & \cellcolor{lightgray!30} & \cellcolor{lightgray!30} & \cellcolor{teal!60} & \cellcolor{teal!60} & \cellcolor{lightgray!30} & \cellcolor{red!40} & \cellcolor{lightgray!30} & \cellcolor{teal!60}
    \\\addlinespace[.25em]
    & T10  
    & $\ast$\cellcolor{teal!60}& \cellcolor{red!40} & 
    \cellcolor{lightgray!30} & \cellcolor{lightgray!30} & 
    \cellcolor{red!40} & \cellcolor{lightgray!30} &
    \cellcolor{red!40} & $\ast$\cellcolor{teal!60} &
    $\ast$\cellcolor{teal!60} & \cellcolor{red!40} & 
    \cellcolor{lightgray!30} & \cellcolor{teal!60}
    \\\addlinespace[.25em]
    & T11  
    & $\ast$\cellcolor{teal!60} & $\ast$\cellcolor{teal!60} & \cellcolor{lightgray!30} & \cellcolor{lightgray!30} & \cellcolor{red!40} & $\ast$\cellcolor{teal!60} & $\ast$\cellcolor{teal!60} & $\ast$\cellcolor{teal!60} & \cellcolor{teal!60} & \cellcolor{teal!60} & \cellcolor{lightgray!30} & \cellcolor{teal!60}
    \\\addlinespace[.25em]
    & T12 
    & \cellcolor{teal!60} & $\ast$\cellcolor{teal!60} & \cellcolor{lightgray!30} & \cellcolor{lightgray!30} & \cellcolor{red!40} & \cellcolor{lightgray!30} & \cellcolor{red!40} & \cellcolor{teal!60} & \cellcolor{teal!60} & \cellcolor{red!40} & \cellcolor{lightgray!30} & \cellcolor{teal!60}

    \\\addlinespace[.25em]
    \bottomrule  
    \addlinespace[.25em]
    && \cellcolor{teal!60} &  \multicolumn{11}{l}{represents successful failure recovery and task completion} \\\addlinespace[.25em]
    && $\ast$\cellcolor{teal!60} &  \multicolumn{11}{l}{represents successful recovery, with no task completion} \\\addlinespace[.25em]
    && \cellcolor{red!40} &  \multicolumn{11}{l}{represents unsuccessful failure recovery} \\
  \end{tabular}
  \caption{Success rate of Ont-RePl replanning with LLMs} 
  \label{tab:results}
\end{table}

The experimental results can be summarized as follows:
1) We demonstrated the capabilities of our rule-based subgoal-verifier, which achieved 100\% on the selected tasks and failures;
2) We demonstrated the capabilities of our ontology-enhanced LLM re-planning pipeline on 90 task-failure pairs, obtaining a good success rate;
3) We show that the safety issues were correctly identified in all tasks and corrected in more than 90\% of cases;
4) We compared with a purely neural online approach consisting of an LLM-based subgoal verifier and an LLM-based re-planner (more details below) and we showed that we significantly outperform both;
5) We demonstrated that with \textsc{Recover} we have a significant reduction of the costs.

\smallskip\noindent
{\bf Baseline model.}~
Most of the failure identification and recovery methods available are designed for an offline setting, and are therefore not suitable for a direct comparison with \textsc{Recover}. One noticeable example is REFLECT~\cite{liu2023reflect}. However, despite the availability of their code, converting it to an online version would necessitate substantial effort.
For this reason, we implemented as baseline a similar approach that is purely LLM-based and uses the same prompting developed in REFLECT.  
The method has two main components: 
1) an LLM-based sub-goal verifier ({\bf LM-SGV}) which is a binary classifier that assesses, at each step of the plan execution, whether an action has been executed successfully or not.
It bases its evaluation on the environment's scene-graph before and after the action, the audio detected, and the list of available objects in the scene.
2) an LLM-based re-planner ({\bf LM-RePl}) that receives as input the scene-graph of the current state of the scene, the list of available objects, the original plan, and the task goal description. It then generates a recovery plan as output.

\smallskip
We compare with the baseline model in two different settings: 
1) Task-failure pairs where \textsc{Recover} successfully identified and corrected the failure (first two colored lines in Table~\ref{tab:results_LM});
2) Task-failure pairs where \textsc{Recover} successfully identified the failure but failed to correct it (third and fourth colored lines in Table~\ref{tab:results_LM}).

\subsection{Sub-goal verification}
Our rule-based sub-goal verifier successfully detected failures in 100\% of the analyzed cases. 
Conversely, the LLM-based sub-goal verifier (LLM-SGV, introduced above) achieves an accuracy of only approximately 50\% in both the analyzed scenarios (task-failure pairs where \textsc{Recover} successfully identified and corrected the failure, and task-failure pairs where \textsc{Recover} successfully identified the failure but failed to correct it). This implies that the failure was correctly identified at the correct execution step in only half of the cases.
The results are provided in Table~\ref{tab:results_LM} under the category LLM-SGV.

\subsection{Task re-planning} 
The failure recovery results for \textsc{Recover} are shown in Table~\ref{tab:results}. The table is divided into two parts: the first block shows the results for easy tasks with at most 20 steps; and the second block shows the results for complex tasks with more than 20 steps.

The recovery rate is around 70\% and it is consistent between easy tasks and complex tasks. However, the completion rate (whether the task is completed successfully after the recovery) is higher for easy tasks (59\%) compared to complex tasks (33\%)
This is to be expected, since with a longer plan, the probability of the LLM to induce an error in the planning increases.

It is evident that certain failures are more straightforward to recover from, such as \emph{MissingNavigationFailure}, while others, such as  \emph{OccupiedByLiquidFailure}, pose greater challenges. 
One contributing factor is the bias inherent in the LLM: for example, the LLM often generates re-plans assuming a robot with two arms, whereas the robot in the AI2Thor simulator is equipped with only one arm.
As a result, the recovery plan may become infeasible because the robot cannot execute actions involving two objects simultaneously; instead, it must perform actions with one object at a time, sequentially.
This discrepancy is particularly noticeable in tasks involving 2 objects that normally would be manipulated simultaneously (e.g., slicing food or pouring liquid). This issue persists even when explicitly specifying in the prompt that the robot has only one arm.

The results of the comparison with the baseline planner (LLM-RePl, introduced above) over easy tasks (when possible) are shown in Table~\ref{tab:results_LM}.  
We can see that in the first block of the table (containing tasks-failure pairs that \textsc{Recover} successfully identified and corrected), LLM-RePl was able to re-plan correctly only 18\% of cases.
In the second block (containing tasks-failure pairs that \textsc{Recover} successfully identified but failed to correct), LLM-RePl is able to re-plan correctly only 11\%. 
We thus demonstrated that our method outperform significantly a pure LLM-based re-planner.

\begin{table}[t]
  \centering
  \begin{tabular}{lKKKKKKKKKKKK}
    \toprule
    & \multicolumn{12}{c}{Failures} \\
    \cmidrule{2-13}
     & \phantom{1}1 & \phantom{1}2 & \phantom{1}3 & \phantom{1}4 & \phantom{1}5 & \phantom{1}6 & \phantom{1}7 & \phantom{1}8 & \phantom{1}9 & 10 & 11 & 12  \\ 
    \midrule
    
    \addlinespace[.4em]
    
    &  \multicolumn{12}{c}{ {\bf Task - Failure} pairs that \textsc{Recover} identified and corrected}  \\
    % \cmidrule{2-13}
    
    \addlinespace[.25em]
    
    LLM-SGV & 
    \cellcolor{teal!60} & \cellcolor{red!40} & \cellcolor{teal!60} & \cellcolor{teal!60} & 
    \cellcolor{red!40} & \cellcolor{teal!60} & 
    \cellcolor{red!40} & \cellcolor{red!40} & 
    \cellcolor{red!40} & \cellcolor{teal!60} & 
    \cellcolor{lightgray!30} & \cellcolor{red!40}
    \\
    
    \addlinespace[.25em]
    
    LLM-RePl & 
    \cellcolor{red!40} & \cellcolor{red!40} & 
    \cellcolor{red!40} & \cellcolor{red!40} & 
    \cellcolor{teal!60} & \cellcolor{red!40} & 
    \cellcolor{red!40} & \cellcolor{teal!60} & 
    \cellcolor{red!40} & \cellcolor{red!40} & 
    \cellcolor{lightgray!30} & \cellcolor{red!40}
    \\
    \addlinespace[.25em]
    Task $\rightarrow$ & T6 & T7 & T3 & T2 & T4 & T6 & T9 & T5 & T8 & T3 & -- & T5  \\  
    \addlinespace[.25em]
    
\midrule

\addlinespace[.4em]

    &  \multicolumn{12}{c}{ {\bf Task - Failure} pairs that \textsc{Recover} identified but failed to correct} \\        
    \addlinespace[.25em]
 
    LLM-SGV & 
    \cellcolor{teal!60} & \cellcolor{red!40} & 
    \cellcolor{lightgray!30} & \cellcolor{red!40} & 
    \cellcolor{red!40} & \cellcolor{lightgray!30} & 
    \cellcolor{teal!60} & \cellcolor{teal!60} & 
    \cellcolor{red!40} & \cellcolor{teal!60} & 
    \cellcolor{red!40} & \cellcolor{teal!60}
    
    \\\addlinespace[.25em]
    LLM-RePl & 
    \cellcolor{teal!60} & \cellcolor{red!40} & 
    \cellcolor{lightgray!30} & \cellcolor{red!40} & 
    \cellcolor{red!40} & \cellcolor{lightgray!30} & 
    \cellcolor{red!40} & \cellcolor{red!40} & 
    NA$^{\diamond}$\cellcolor{lightgray!30}& \cellcolor{red!40} & 
    \cellcolor{red!40} & \cellcolor{red!40}
    
    \\
    \addlinespace[.25em]
    Task $\rightarrow$ & T1 & T10 & -- & T7 & T2 & -- & T3 & T4 & T10 & T1 & T2 & T1  \\  
    \addlinespace[.25em]
      \bottomrule  
        \addlinespace[.25em]
  & \cellcolor{teal!60} &  \multicolumn{11}{l}{represents successful failure recovery and task completion} \\\addlinespace[.25em]
  &  \cellcolor{red!40} &  \multicolumn{11}{l}{represents unsuccessful failure recovery}\\ 
  &  & \multicolumn{11}{l}{$^{\diamond}$stopped since reached limit of 5 US\$}
  \\
  \end{tabular}%}
    \caption{Success rate of LLM-SGV and LLM-RePl for easy tasks}
  \label{tab:results_LM}
\end{table}
\begin{table*}[!ht]
  \centering\footnotesize
  \begin{tabular}{lKKKKKKKKKKKK}
    \toprule
    & \multicolumn{12}{c}{Failures} \\
    \cmidrule{2-13}
     & \phantom{1}1 & \phantom{1}2 & \phantom{1}3 & \phantom{1}4 & \phantom{1}5 & \phantom{1}6 & \phantom{1}7 & \phantom{1}8 & \phantom{1}9 & 10 & 11 & 12 \\ 
    \midrule    
    \textsc{Recover} &
    \bf 0.08 & \bf 0.13 & \bf 0.21 & 0.28 & 
    \bf0.15 & \bf0.09 & \bf0.08 & \bf0.11 & 
    \bf0.08 & \bf0.19 & -- & \bf 0.16  \\
    LLM-based &
    0.14 & 0.16 & 0.91 & \bf 0.16 & 
    1.04 & 0.82 & 0.21 &  0.16 & 
    1.00 & 0.51 &  -- & 0.36  \\
    Task $\rightarrow$ & T6 & T7 & T3 & T2 & T4 & T6 & T9 & T5 & T8 & T3 & -- & T5  \\  

\midrule
    \textsc{Recover} &
    \bf0.13 & \bf0.14 & -- & \bf0.06 & 
    \bf0.08 & -- & \bf0.18 & \bf0.16 & 
    \bf0.07 & \bf0.09 & \bf0.08 & \bf0.06  \\
    LLM-based &
    0.27 & 2.13 & -- & 0.16 & 
    0.57 & -- & 0.54 & 0.58 & 
    $>$5 & 0.42 & 0.57 & 0.27  \\
    \addlinespace[.25em]
    Task $\rightarrow$ & T1 & T10 & -- & T7 & T2 & -- & T3 & T4 & T10 & T1 & T2 & T1  \\  
      \bottomrule   
  \end{tabular}
    \caption{Cost of \textsc{Recover} vs LLM-based approach (in US\$)}
  \label{tab:cost}
\end{table*}

\subsection{Additional Results}

{\bf Safety.}~ 
Our method identifies 100\% of the safety issues that occurred during task executions, and it recovers successfully from 93\% of them\footnote{This percentage considers only the  successful tasks. When failed tasks are also considered, the recovery rate is 86\%.}, restoring the safety of the environment. 
The LLM-based method instead only identifies and corrects 31\% of the safety issues in the scenes analyzed.

\smallskip\noindent
{\bf Cost.}~ 
As indicated in Table~\ref{tab:cost} , our approach demonstrates a significantly greater cost-effectiveness compared to an LLM-based method,  resulting in a noticeable reduction in monetary costs. 
It is important to note that the value reported for the LLM-based method serves as a conservative estimate and may underestimate the actual costs.
In our experiments, this value is heavily influenced by the step in the plan at which the failure occurs.
The later the failure arises, the higher the associated cost, as the LLM-based sub-goal verifier queries the LLM at each step until a failure is detected.
If the LLM fails to identify the failure, it continues to query the LLM at each subsequent step. 
This situation is exemplified by task T10 with failure 9, where the cost exceeded our predefined threshold of 5 US\$.

%%%%%%%%%%%%%%%%%%%%%%%%%%%%%%%%%%%%%%%%%%%%%%%%%%%%%%%%%%%%%%%%%%%

\section{Conclusions and Future Work}

We presented \textsc{Recover}, a neuro-symbolic framework designed for online failure detection and recovery. 
Our approach exhibits good performance, surpassing a LLM-based baseline method. 
Moreover, we demonstrated it to be cost-effective. 
Finally, given its personalization capability, it is able to identify and correct safety issues in an environment.

In future extensions, we plan to broaden the scope of this work by exploring simulator scenarios with human interaction. This entails studying how ontology-based failure detection and recovery mechanisms function when humans are actively engaged in the perception-action loop or when specialized knowledge has crucial impact. 
A deeper integration of LLMs with ontologies \cite{pan2024unifying} within a hybrid framework could significantly augment both reasoning capabilities and re-planning quality.
Finally, further refinements in prompt engineering would undoubtedly enhance the system's performance.
%%%%%%%%%%%%%%%%%%%%%%%%%%%%%%%%%%%%%%%%%%%%%%%%%%%%%%%%%%%%%%%%%%%

\bibliographystyle{plain}

\begin{thebibliography}{10}

\bibitem{swi_prolog}
{SWI}-prolog.
\newblock {https://www.swi-prolog.org}.
\newblock Version: 8.3.3.

\bibitem{ahn2022can}
Michael Ahn, Anthony Brohan, Noah Brown, Yevgen Chebotar, Omar Cortes, Byron David, Chelsea Finn, Chuyuan Fu, Keerthana Gopalakrishnan, Karol Hausman, et~al.
\newblock Do as i can, not as i say: Grounding language in robotic affordances.
\newblock {\em arXiv preprint arXiv:2204.01691}, 2022.

\bibitem{chatzilygeroudis2019survey}
Konstantinos Chatzilygeroudis, Vassilis Vassiliades, Freek Stulp, Sylvain Calinon, and Jean-Baptiste Mouret.
\newblock A survey on policy search algorithms for learning robot controllers in a handful of trials.
\newblock {\em IEEE Transactions on Robotics}, 36(2):328--347, 2019.

\bibitem{das2021semantic}
Devleena Das and Sonia Chernova.
\newblock Semantic-based explainable ai: Leveraging semantic scene graphs and pairwise ranking to explain robot failures.
\newblock In {\em 2021 IEEE/RSJ International Conference on Intelligent Robots and Systems (IROS)}, pages 3034--3041. IEEE, 2021.

\bibitem{diab2019ontology}
Mohammed Diab, Mihai Pomarlan, Daniel Be{\ss}ler, Aliakbar Akbari, Jan Rosell, John Bateman, and Michael Beetz.
\newblock An ontology for failure interpretation in automated planning and execution.
\newblock In {\em Iberian Robotics conference}, pages 381--390. Springer, 2019.

\bibitem{golovneva2022roscoe}
Olga Golovneva, Moya~Peng Chen, Spencer Poff, Martin Corredor, Luke Zettlemoyer, Maryam Fazel-Zarandi, and Asli Celikyilmaz.
\newblock Roscoe: A suite of metrics for scoring step-by-step reasoning.
\newblock In {\em The Eleventh International Conference on Learning Representations}, 2022.

\bibitem{guzhov2022audioclip}
Andrey Guzhov, Federico Raue, J{\"o}rn Hees, and Andreas Dengel.
\newblock Audioclip: Extending clip to image, text and audio.
\newblock In {\em ICASSP 2022-2022 IEEE International Conference on Acoustics, Speech and Signal Processing (ICASSP)}, pages 976--980. IEEE, 2022.

\bibitem{huang2022language}
Wenlong Huang, Pieter Abbeel, Deepak Pathak, and Igor Mordatch.
\newblock Language models as zero-shot planners: Extracting actionable knowledge for embodied agents.
\newblock In {\em International Conference on Machine Learning}, pages 9118--9147. PMLR, 2022.

\bibitem{jin2024robotgpt}
Yixiang Jin, Dingzhe Li, A~Yong, Jun Shi, Peng Hao, Fuchun Sun, Jianwei Zhang, and Bin Fang.
\newblock Robotgpt: Robot manipulation learning from chatgpt.
\newblock {\em IEEE Robotics and Automation Letters}, 2024.

\bibitem{sgg}
Boris Knyazev, Harm de~Vries, Catalina Cangea, Graham~W. Taylor, Aaron~C. Courville, and Eugene Belilovsky.
\newblock Graph density-aware losses for novel compositions in scene graph generation.
\newblock In {\em BMVC}, 2020.

\bibitem{ai2thor}
Eric Kolve, Roozbeh Mottaghi, Winson Han, Eli VanderBilt, Luca Weihs, Alvaro Herrasti, Daniel Gordon, Yuke Zhu, Abhinav Gupta, and Ali Farhadi.
\newblock {AI2-THOR: An Interactive 3D Environment for Visual AI}.
\newblock {\em arXiv}, 2017.

\bibitem{lamy2017owlready}
Jean-Baptiste Lamy.
\newblock Owlready: Ontology-oriented programming in python with automatic classification and high level constructs for biomedical ontologies.
\newblock {\em Artificial intelligence in medicine}, 80:11--28, 2017.

\bibitem{li2023semantically}
Wenhao Li, Dan Qiao, Baoxiang Wang, Xiangfeng Wang, Bo~Jin, and Hongyuan Zha.
\newblock Semantically aligned task decomposition in multi-agent reinforcement learning.
\newblock {\em arXiv preprint arXiv:2305.10865}, 2023.

\bibitem{liu2023reflect}
Zeyi Liu, Arpit Bahety, and Shuran Song.
\newblock {REFLECT}: Summarizing robot experiences for failure explanation and correction.
\newblock In {\em 7th Annual Conference on Robot Learning}, 2023.

\bibitem{nenov2015rdfox}
Yavor Nenov, Robert Piro, Boris Motik, Ian Horrocks, Zhe Wu, and Jay Banerjee.
\newblock Rdfox: A highly-scalable rdf store.
\newblock In {\em The Semantic Web-ISWC 2015: 14th International Semantic Web Conference, Bethlehem, PA, USA, October 11-15, 2015, Proceedings, Part II 14}, pages 3--20. Springer, 2015.

\bibitem{pan2024unifying}
Shirui Pan, Linhao Luo, Yufei Wang, Chen Chen, Jiapu Wang, and Xindong Wu.
\newblock Unifying large language models and knowledge graphs: A roadmap.
\newblock {\em IEEE Transactions on Knowledge \& Data Engineering}, (01):1--20, 2024.

\bibitem{ren2023robots}
Allen~Z Ren, Anushri Dixit, Alexandra Bodrova, Sumeet Singh, Stephen Tu, Noah Brown, Peng Xu, Leila Takayama, Fei Xia, Jake Varley, et~al.
\newblock Robots that ask for help: Uncertainty alignment for large language model planners.
\newblock In {\em 7th Annual Conference on Robot Learning}, 2023.

\bibitem{schick2023toolformer}
Timo Schick, Jane Dwivedi-Yu, Roberto Dessi, Roberta Raileanu, Maria Lomeli, Eric Hambro, Luke Zettlemoyer, Nicola Cancedda, and Thomas Scialom.
\newblock Toolformer: Language models can teach themselves to use tools.
\newblock In {\em Thirty-seventh Conference on Neural Information Processing Systems}, 2023.

\bibitem{shinn2023reflexion}
Noah Shinn, Federico Cassano, Edward Berman, Ashwin Gopinath, Karthik Narasimhan, and Shunyu Yao.
\newblock Reflexion: Language agents with verbal reinforcement learning.
\newblock In {\em arXiv/2303.11366}, 2023.

\bibitem{vemprala2023chatgpt}
Sai Vemprala, Rogerio Bonatti, Arthur Bucker, and Ashish Kapoor.
\newblock Chatgpt for robotics: Design principles and model abilities.
\newblock {\em arXiv preprint arXiv:2306.17582}, 2023.

\bibitem{wang2023learning}
Danqing Wang and Lei Li.
\newblock Learning from mistakes via cooperative study assistant for large language models.
\newblock In {\em The 2023 Conference on Empirical Methods in Natural Language Processing}, 2023.

\bibitem{narangself}
Xuezhi Wang, Jason Wei, Dale Schuurmans, Quoc~V Le, Ed~H. Chi, Sharan Narang, Aakanksha Chowdhery, and Denny Zhou.
\newblock Self-consistency improves chain of thought reasoning in language models.
\newblock In {\em The Eleventh International Conference on Learning Representations}, 2023.

\bibitem{wei2022emergent}
Jason Wei, Yi~Tay, Rishi Bommasani, Colin Raffel, Barret Zoph, Sebastian Borgeaud, Dani Yogatama, Maarten Bosma, Denny Zhou, Donald Metzler, Ed~H. Chi, Tatsunori Hashimoto, Oriol Vinyals, Percy Liang, Jeff Dean, and William Fedus.
\newblock Emergent abilities of large language models.
\newblock {\em Transactions on Machine Learning Research}, 2022.

\bibitem{chain_of_thoughts}
Jason Wei, Xuezhi Wang, Dale Schuurmans, Maarten Bosma, Ed~H. Chi, Quoc Le, and Denny Zhou.
\newblock Chain of thought prompting elicits reasoning in large language models.
\newblock {\em CoRR}, abs/2201.11903, 2022.

\bibitem{wu2023tidybot}
Jimmy Wu, Rika Antonova, Adam Kan, Marion Lepert, Andy Zeng, Shuran Song, Jeannette Bohg, Szymon Rusinkiewicz, and Thomas Funkhouser.
\newblock Tidybot: Personalized robot assistance with large language models.
\newblock {\em Autonomous Robots}, 2023.

\bibitem{yao2020calm}
Shunyu Yao, Rohan Rao, Matthew Hausknecht, and Karthik Narasimhan.
\newblock Keep calm and explore: Language models for action generation in text-based games.
\newblock In {\em Empirical Methods in Natural Language Processing (EMNLP)}, 2020.

\bibitem{yao2022react}
Shunyu Yao, Jeffrey Zhao, Dian Yu, Nan Du, Izhak Shafran, Karthik~R Narasimhan, and Yuan Cao.
\newblock React: Synergizing reasoning and acting in language models.
\newblock In {\em The Eleventh International Conference on Learning Representations}, 2023.

\bibitem{yao2024retroformer}
Weiran Yao, Shelby Heinecke, Juan~Carlos Niebles, Zhiwei Liu, Yihao Feng, Le~Xue, Rithesh~R N, Zeyuan Chen, Jianguo Zhang, Devansh Arpit, Ran Xu, Phil~L Mui, Huan Wang, Caiming Xiong, and Silvio Savarese.
\newblock Retroformer: Retrospective large language agents with policy gradient optimization.
\newblock In {\em The Twelfth International Conference on Learning Representations}, 2024.

\bibitem{zhao2023expel}
Andrew Zhao, Daniel Huang, Quentin Xu, Matthieu Lin, Yong-Jin Liu, and Gao Huang.
\newblock Expel: Llm agents are experiential learners.
\newblock In {\em arXiv/2308.10144}, 2023.

\end{thebibliography}

\end{document}